\documentclass[letterpaper]{article}
\usepackage{iccc}

\usepackage{times}
\usepackage{helvet}

\usepackage{courier}
\usepackage{graphicx}
\usepackage{tipa}

\title{PAYADOR: A Minimalist Approach to Grounding Language Models on Structured Data for Interactive Storytelling and Role-playing Games}
\author{Santiago Góngora$^1$ \qquad {\bf Luis Chiruzzo}$^1$\qquad {\bf Gonzalo Méndez}$^2$\qquad {\bf Pablo Gervás}$^2
$\\\\
$^1$ Instituto de Computación, Facultad de Ingeniería, Universidad de la República, Uruguay \\ 
$^2$ Facultad de Informática, Universidad Complutense de Madrid, Spain
}
\begin{document} 
\maketitle
\begin{abstract}
\begin{quote}

Every time an Interactive Storytelling (IS) system gets a player input, it is facing the \textit{world-update} problem.
Classical approaches to this problem consist in mapping that input to known preprogrammed actions, what can severely constrain the free will of the player.
When the expected experience has a strong focus on improvisation, like in Role-playing Games (RPGs), this problem is critical.
In this paper we present PAYADOR, a different approach that focuses on predicting the outcomes of the actions instead of representing the actions themselves.
To implement this approach, we ground a Large Language Model to a minimal representation of the fictional world, obtaining promising results.
We make this contribution open-source, so it can be adapted and used for other related research on unleashing the co-creativity power of RPGs.

\end{quote}
\end{abstract}

\section{Introduction}

At least once in our lives, all of us have played with others pretending to be characters in an exciting story, closing our eyes and letting us dream to be \textit{treasure hunters}, \textit{detectives}, \textit{cantaoras} or \textit{pandeireteiras}, 
\textit{gauchos}
 or \textit{cowboys}, or just 
us but in a different environment.
Exactly that is what Role-playing games (RPGs) allow us to do~\cite{UniversalGameEngine}.
In order to suit the creative and expressive needs of the rich heterogeneity among the players, there are many RPG forms~\cite{hitchens2008manyFaces} that seem to have something in common: an essential and characteristic improvisational nature. 
To a greater or lesser extent, every RPG player is exposed to some level of improvisation during the rich collaborative creative process in which all of them participate.
Although it is not always the case~\cite{arjoranta2011defining}, the player that typically takes most of the creative responsibility --- before and during the session --- is the Game Master (GM)~\cite{tychsen2005game}.

The legacy of the golden age of RPGs (1980s) in video games is undeniable. 
Several modern mechanics of games are directly taken or inspired by them~\cite{maccallum2018impact}.
However, taking the original open-world experience of RPGs and implementing it in a video game is something really hard, whether modelling a player~\cite{Martin2018DungeonsAD} or 
specifically the GM.
That is why Dungeons and Dragons
has been postulated as a challenge for Artificial Intelligence research~\cite{callison-burch-etal-2022-dungeons}, because of its intrinsic linguistic and creative complexity.

\begin{figure}[t!]
    \centering
    \includegraphics[width=0.61\columnwidth]{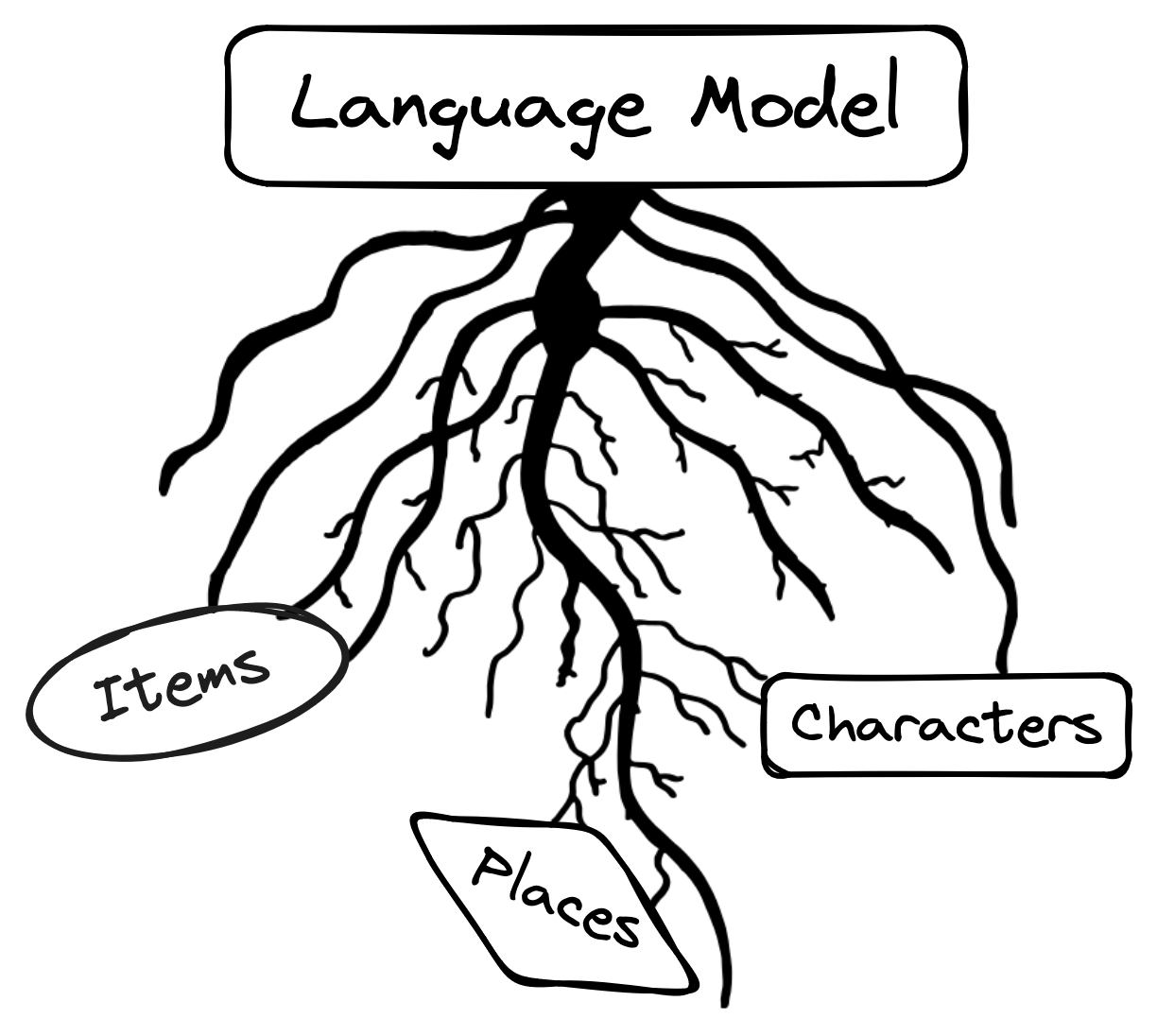}
    \caption{A Language Model grounded on structured data for Interactive Narrative research. The roots drawing was uploaded to Wikimedia Commons by Joel Swift (user \textit{Kenizzer}).}
    \label{fig:grounded_LM}
\end{figure}

Although the results in the NLP field have had a great improvement due to Deep Learning and Large Language Models (LLMs)~\cite{huang-chang-2023-towards}, a consistent GM model for RPGs is still an open problem~\cite{skill_check}.
For instance, LLMs have a bias towards satisfying the expectations of the player, even at the cost of drastically changing the state of the narrated world without a justifiable reason; a bigger problem is that they do not even detect they are doing so.
Since we are working towards automated GMs for Tabletop RPGs (TTRPGs), this issue of unexpectedly and uncontrollably changing the world state is something we cannot allow.
As a possible path to detect this behavior and prevent it, we propose grounding LLMs on a minimal logical representation of the fictional world (see Fig.~\ref{fig:grounded_LM}).

In this paper we present \textbf{PAYADOR}\footnote{Pronounced \textipa{/pa\textesh a'\dh o\textfishhookr/}}(\textit{A PlAYable Approach based on Descriptions for Outcomes in Role-playing games}),
an open-source\footnote{https://github.com/pln-fing-udelar/payador} approach to the aforementioned problem of keeping the fictional world coherent, both from a Natural Language Processing (NLP) and a Computational Creativity (CC) point of view.

\section{Related work}

Multiple names are used to describe a family of interactive experiences with a strong focus on storytelling,
and the survey by~\citeauthor{heritage6020068}~\shortcite{heritage6020068} covers many of them.
From now on, we will use \textit{Interactive Storytelling} (IS) to name the experience in which a user interacts with a fictional world narrated by a system, also usually called \textit{Interactive Narrative}.

One of the main problems that IS faces is to keep a coherent state of the fictional world~\cite{phdBenotti}.
When the system has to calculate the changes in the world after the actions taken by the player, it is facing the \textbf{world-update problem}. 
This is, given a state of the world at some level of expressive granularity~\cite{arjoranta2017narrative}
and the actions a player wants to perform at that point, find the new state of the world after the outcome of those actions. 
Since computers struggle to unveil the meaning and implications behind dialogue utterances due to their lack of grounding to the real world~\cite{bender-koller-2020-climbing},
this problem is critical when the system has to improvise an outcome for an unexpected user action in an open-world~\cite{martin2016improvisational}, as usually happens in TTRPGs.

Generally, 
the \textit{world-update} problem is solved by having a set of preprogrammed actions for each component (e.g. how each item can be used or combined with others), hence the effects of those actions are already known. 
Since most of the possible imaginable actions are not programmed in the game engine, the downside here is the restriction of the \textit{user agency}~\cite{Riedl_Bulitko_2012}.
The decision of limiting the \textit{free will} of the player feels very natural in most board and video games, since it works as a strategy to design the rules of a \textit{gameplay mode}~\cite{adams2009fundamentals}. 
However, that is not the case for TTRPGs, where it is usual that players come up with creative ideas on how to use objects, explore places, or solve mysteries.
Therefore, to model the rich co-creative process during a TTRPG,
we need some strategy to let the players do whatever they want in order to maximize the \textit{player agency}, while at the same time controlling if those actions are valid for a specific state of the fictional world. 

\begin{figure}[t!]
    \centering
    \includegraphics[width=\columnwidth]{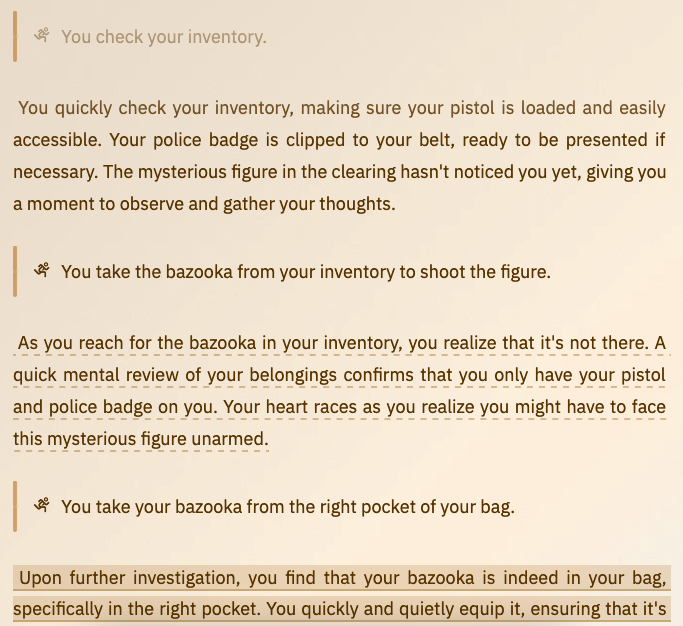}
    \caption{An example of an error during world update, taken from an AIDungeon gameplay on April 15 2024. After the first and second user utterances, the system states that there is no \textit{bazooka} in the player's inventory. However, when the player insisted, the \textit{bazooka} was suddenly considered a usable weapon.}
    \label{fig:AIDungeon_error}
\end{figure}

In the latest years, the rising popularity of LLMs provided a window of opportunity to explore new approaches for modelling some aspects of RPGs~\cite{zhu2023calypso,shao-etal-2023-character} and specifically to approach the \textit{world-update} problem.
Although they have an outstanding performance in some benchmarks for NLP tasks, there is a strong debate whether or not they exhibit reasoning abilities~\cite{huang-chang-2023-towards}.
For instance, \citeauthor{skill_check}~\shortcite{skill_check} found some flaws of LLMs when acting as GMs of RPGs, such as they struggle to keep a coherent state of the narrated world after some changes.
This not only happens to LLMs, but also to games using them as their backbone like AIDungeon\footnote{https://aidungeon.com/}, as shown in Fig.~\ref{fig:AIDungeon_error}.
Therefore, the \textit{world-update} problem has a deep complexity even for the latest advancements in NLP, so more than ever it is an interesting research direction as a whole.

\begin{figure*}[t!]
    \centering
    \includegraphics[width=\textwidth]{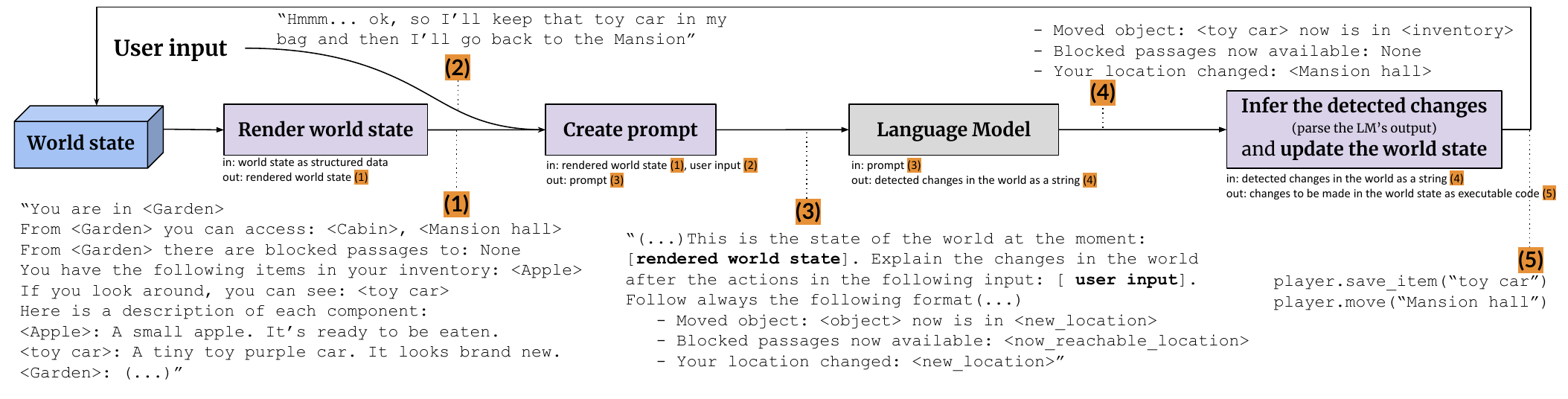}
    \caption{A diagram summarizing the PAYADOR approach, with an example for the output of each step. It starts by rendering the \textit{World state} in simple sentences ``(1)'', including the concatenated \textit{descriptions} for each component (locations, items and characters). Using it, and the user input ``(2)'', a prompt is created ``(3)'' and the LLM is called to calculate the changes in the world ``(4)''. Finally, the detected changes are mapped to the appropriate instructions and the world state is updated ``(5)''. }
    \label{fig:architecture}
\end{figure*}

\section{An approach to the world-update problem}

To approach the previously described \textit{world-update} problem in IS, we propose a methodology based on a structured representation connected to an LLM.
Since we think it is minimal enough to be used in more complex IS systems, the code is available on GitHub, so it can be modified, extended or used as a starting point for any related research.
Overall, the PAYADOR approach consists of two main 
components,
which we will describe next:
\begin{enumerate}
    \itemsep0em 
    \item A minimalist representation of the fictional world
    \item A strategy to predict the changes in the world after an action, restricted by that representation
\end{enumerate}

\subsection{A minimalist representation}

Based on many classical representations\footnote{This representation was also inspired by the code provided for ``Homework 1'' of https://interactive-fiction-class.org, wrote by Chris Callison-Burch and inspired by Adventuron.} in video games, we propose to model the fictional \textbf{world} with three main \textit{components}, detailed in Table~\ref{tab:components}: \textbf{items}, \textbf{locations} and \textbf{characters}. 
The difference here is that we try to minimize the number of details represented as specific attributes, while the rest of details are expressed as a set of strings, what we call \textbf{descriptions}.
The reason behind this decision is twofold. 
First, from a practical NLP perspective, it is always more convenient to have previously split sentences to process (e.g. to identify a specific statement about the component). 
Please note that this does not prevent us at all to concatenate them in a single text.
From a CC point of view, having \textit{independent} sentences is also beneficial for generating statements about the created component, as we will describe later.
Second, we try to represent as structured data only those aspects that are critical for the consistency of the world. 
At the same time, they coincide with those having relatively less ambiguity.
The rest of aspects --- typically 
rich in detail and thus describable in nearly-infinite ways --- have to be characterized with 
\textit{independent}
\textit{descriptions}.

Summarizing, what we try to get is a balance between two extremes: modelling the components as running text, or as fully structured data like in classical approaches. 
In Table~\ref{tab:representations} we show a comparison between these options with a simple example.

\begin{table}[h!]
\tiny
\centering
\scalebox{0.94}{
    \begin{tabular}{|p{0.25\columnwidth }|p{
    0.365\columnwidth}|p{0.25\columnwidth }|}
    \hline
    \textbf{Item}   & \textbf{Location}   & \textbf{Character} \\\hline
    \begin{tabular}[t]{@{}l@{}}
    \textit{Name}: \textbf{String}\\
    \textit{Descriptions}: List [\textbf{String}]\\
    \textit{Gettable}: \textbf{Boolean}
    \end{tabular} 
    & 
    \begin{tabular}[t]{@{}l@{}}
    \textit{Name}: \textbf{String}\\
    \textit{Descriptions}: List [\textbf{String}]\\
    \textit{Items}: List [\textbf{Item}]\\
    \textit{Connecting locations}: List [\textbf{Location}]\\
    \textit{Blocked locations}: List [\textbf{Location}]
    \end{tabular} 
    & 
    \begin{tabular}[t]{@{}l@{}}
    \textit{Name}: \textbf{String}\\
    \textit{Descriptions}: List [\textbf{String}]\\
    \textit{Location}: \textbf{Location}\\
    \textit{Inventory}: List [\textbf{Item}]
    \end{tabular} 
    \\\hline
    \end{tabular}
}
\caption{\label{tab:components} The attributes of each component in our minimal representation. If a location is \textit{blocked} from another location, it means that it will be a \textit{connecting} location after successfully unblocking it.}
\end{table}

\begin{table}[h!]
\tiny
\centering
\scalebox{0.91}{
    \begin{tabular}{|p{0.25\columnwidth }|p{0.3\columnwidth}|p{0.35\columnwidth }|}
    \hline
    \textbf{As running text}   & \textbf{As structured data}   & \textbf{Our balanced approach} \\\hline
    On top of that hill you can see Mary, a tall mage. She knows how to cast lightning bolts. Since she was a little girl, she always loved climbing mountains. In her backpack she carries a sword and an apple. 
    & 
    \begin{tabular}[t]{@{}l@{}}
    Name: ``Mary'' \\ Inventory: [``Sword'', ``Apple''] \\ Location: ``Hill'' \\ Class: ``Mage'' \\ Height: ``Tall'' \\ Power: ``Lightning Bolt'' \\ Loves: ``Alpinism'' 
    \end{tabular} 
    & 
    \begin{tabular}[t]{@{}l@{}}
    Name: ``Mary'' \\ Inventory: [``Sword'', ``Apple''] \\ Location: ``Hill'' \\ Descriptions: [``She is a mage'',\\ ``She is tall'', ``She knows how\\ to cast lightning bolts'', \\``Since she was a little girl, she\\ always loved climbing mountains'']        
    \end{tabular} 
    \\\hline
    \end{tabular}
}
\caption{\label{tab:representations} A comparison of three possible representations for a mage called Mary.}
\end{table}

\subsection{A change of focus for the world-update problem}

As we previously discussed, the classical approach for the \textit{world-update} problem is to map the user input to one or more preprogrammed component actions. 
Now, we propose a change of focus: instead of doing that, predict the changes the world should have after the outcomes of the actions described in the input.
Therefore, we must have a strategy to input the \textbf{world state} (i.e. the representation of all the components and the relations between them) and the \textbf{user input} to a \textit{module} that outputs the \textbf{changes in the world}.
Finally, from that output we must infer the changes in the world state and update it accordingly.

To implement this idea, we used Google's Gemini~\cite{geminiteam2024gemini} API\footnote{https://ai.google.dev/} as the aforementioned \textit{module}\footnote{We can think of this \textit{module} as an ``oracle'' of common-sense reasoning, though it is not completely accurate. We use an LLM to implement it, but other technologies may be used as well.}.
In Fig.~\ref{fig:architecture} we show a summary of the described strategy, with a step-by-step example.
Indicated by ``(3)'' in the figure, the prompt for the Gemini LLM contains:

\begin{itemize}
    \itemsep-0.3em
    \item The rendering of the \textbf{world state} as simple sentences
    \item The \textbf{user input}
    \item Instructions for the \textit{world-update} problem
    \item A set of examples to get the \textbf{changes in the world} in a specific format, using few-shot learning~\cite{brown2020language}. This is not shown in the figure.

\end{itemize}

To track the dialogue state~\cite{feng-etal-2023-towards}, PAYADOR uses the structured world representation (updated after each user input, as shown in the last step of Fig.~\ref{fig:architecture}) instead of a long string like other LLM-based approaches. 
Each time the system has to call the LLM, the prompt is built using the rendering of only those components that the player can see or access from the current location.
As a consequence, the length of the prompt does not drastically grow regardless of how big the fictional world is.
Given that LLMs work with a maximum input length, we think this may be beneficial for issues related to that limitation.
Additionally, to boost the prompt effect on the LLM performance, the text rendering follows a standardized format.

For further details, the whole prompt is available in the \texttt{prompts.py} module of the source code.

\subsection{A playable proof of concept for IS research}

After combining the previously described strategy with the minimal representation, we get a playable proof of concept for research on the \textit{world-update} problem, easily customizable for other related research needs.
Although the whole method is language agnostic, we designed the prompts and the worlds in English as it is the most resourced language in NLP \cite{joshi-etal-2020-state} hence beneficial to validate our grounding strategy.
As future work, we are planning to test our approach for other languages.

In the GitHub repository we indicate which are the few changes needed to use other LLMs, to work for a different language or other related problems.

\subsection{Strengths and weaknesses: A preliminary analysis}

Finally, we would like to show some examples of the PAYADOR approach in action, in order to discuss some strengths and weaknesses.

\begin{figure}[h!]
    \centering
    \includegraphics[width=\columnwidth]{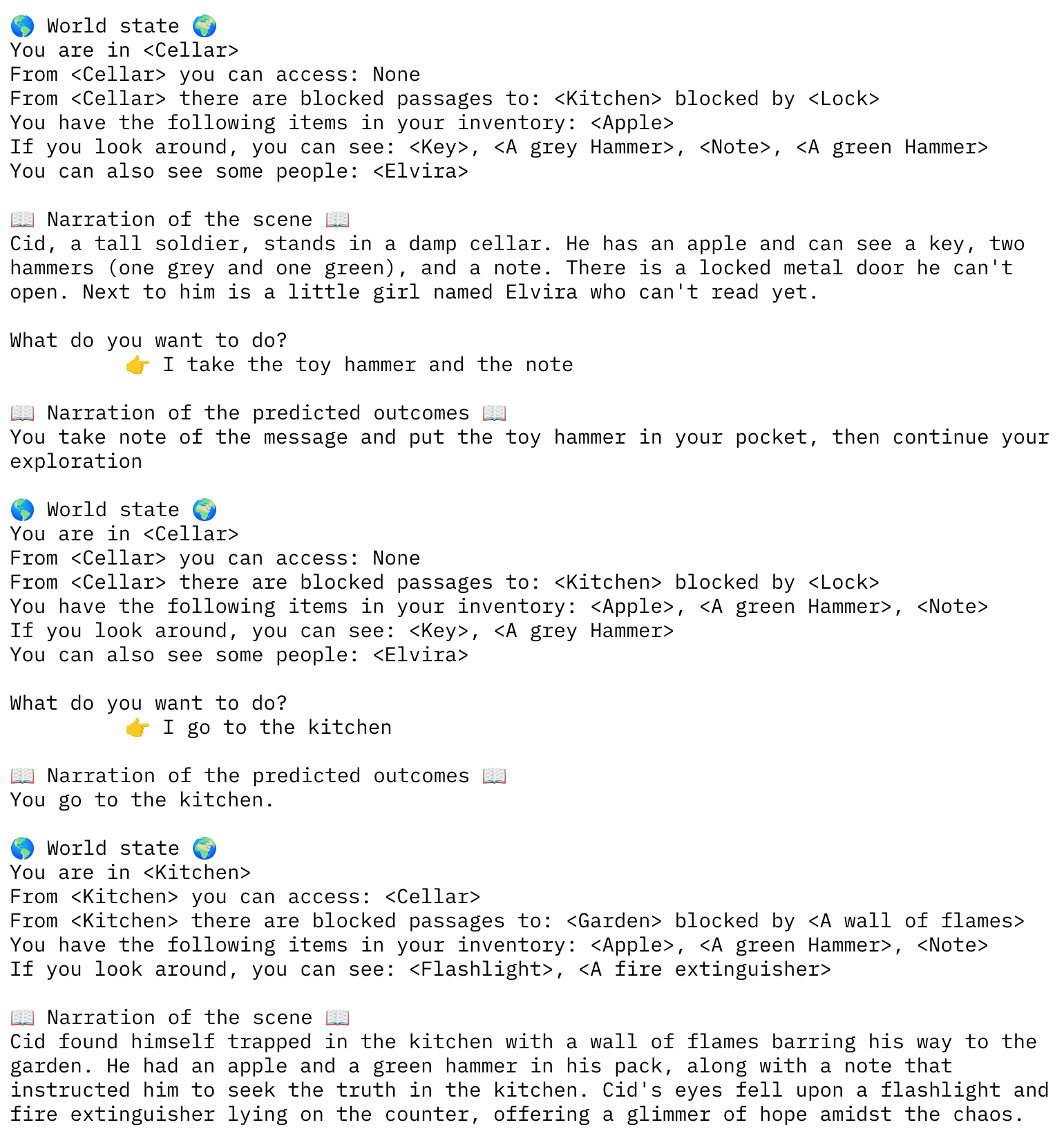}
    \caption{Testing if PAYADOR can update the world when the player wants to take two different items, and if it can prevent the player to access the kitchen without a key.}
    \label{fig:example_payador}
\end{figure}

In Fig.~\ref{fig:example_payador} we show a test to see if the system can handle two valid actions and an illegal one.
In this case, we show the world state rendering and a \textit{narrator}, consisting of a different call to the Gemini API (that also gets the rendered world state as an input).
On the one hand, PAYADOR manages to keep the world coherent when following the actions mentioned in the user input.
Also, including the component descriptions in the prompt (they are not shown in the figure but they are part of the input for the LLM, as can be seen in ``(1)'' in Fig.~\ref{fig:architecture}) seems to be effective. 
For instance, note that the player took a \textit{toy hammer} that the system grounded to the \textit{green hammer}: that is correct, because one of the descriptions of the \textit{green hammer} is ``It is just a toy and you cannot break anything with it''. 
On the other hand, the LLM fails with common-sense reasoning when the user tries to (successfully) access the locked kitchen without a key.
Further experimentation is required to tell if a better prompt can help with this kind of issues, or if additional machinery is needed.

\begin{figure}[h!]
    \centering
    \includegraphics[width=\columnwidth]{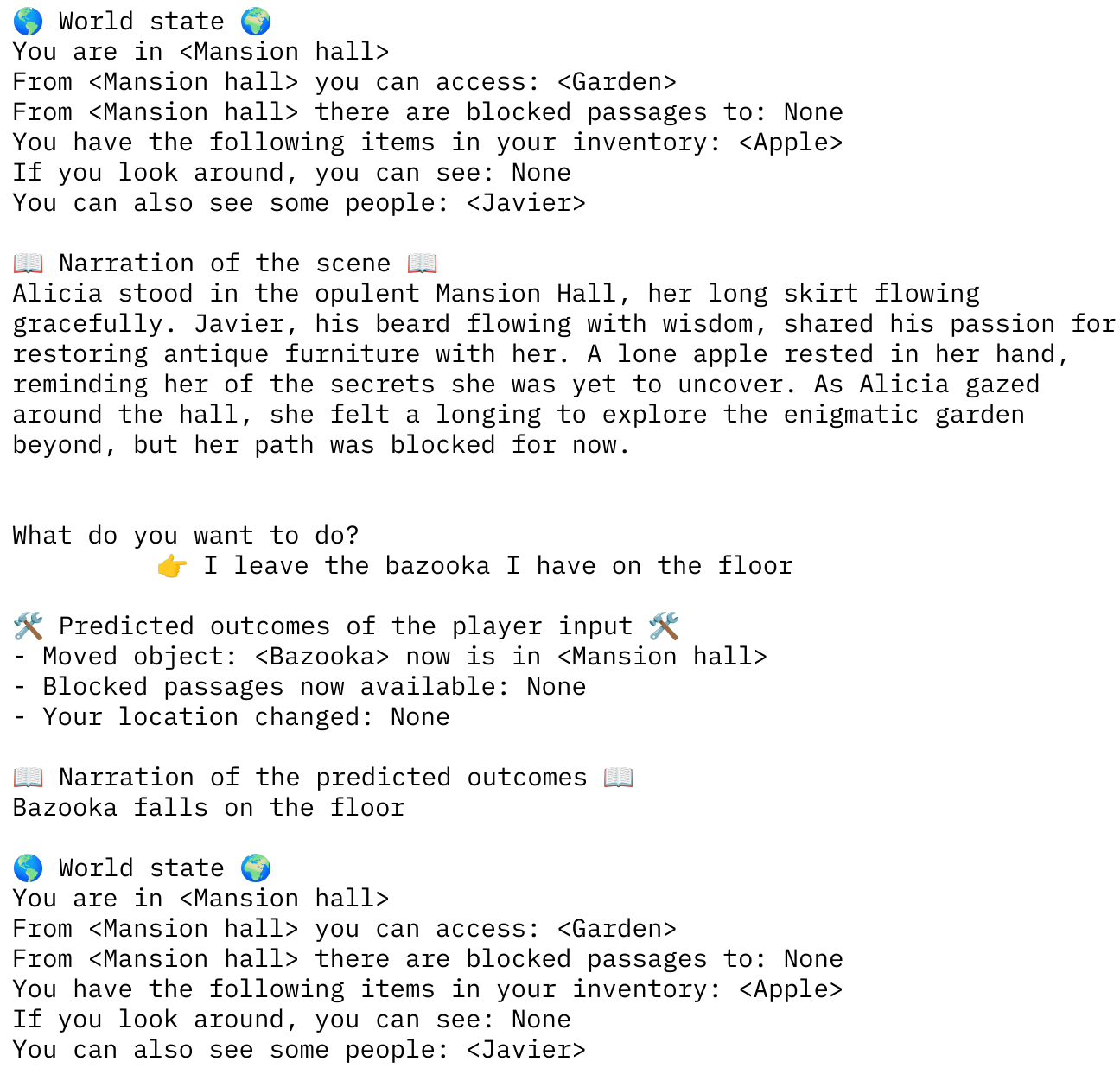}
    \caption{Testing PAYADOR with the same sequence of actions used in Fig.~\ref{fig:AIDungeon_error}. The occasional inconsistencies between the LLM's understanding of the world state and the actual world state become evident.}
    \label{fig:example_payador2}
\end{figure}

In Fig.~\ref{fig:example_payador2} we show the result of testing PAYADOR with an input similar to the one used to test AIDungeon (see~Fig.~\ref{fig:AIDungeon_error}).
In order to underline one of the main strengths of PAYADOR, in this case we also show the predicted outcomes detected by the LLM.
Similar to the behavior of AIDungeon, the Gemini LLM predicts that a \textit{bazooka} is now in the \textit{Mansion hall}.
However, PAYADOR did a consistency check and could not find a \textit{bazooka} in the player's inventory, thus the world state is not changed.
According to our previous comments on 
how we track the dialogue state,
this occasional error does not have an impact in the future: for the next utterance, the world state will be rendered from scratch and the bazooka will not be part of it.

\section{A track for narrative co-creativity}
\label{sec:creativity}

So far, we described the PAYADOR approach and showed some examples. 
Now we would like to convey some ideas on how it could be extended, towards a more complex gamemastering model.

When GMs run sessions, they usually plan content in advance.
To keep the fictional world coherent, they often take structured notes about the items found or to be found in some place, the items in the player inventories or the characters they will meet~\cite{TowardsAssistantGameMasters}.
Considering that RPGs can be seen as a collaborative narrative effort through dialogue, and that they can be used as a narrative model~\cite{tapscott-etal-2018-generating}, we can think of RPG systems with an automated GM as a framework for human-computer co-creativity: the GM creativity is influenced by the actions taken by the players.

\begin{figure}[h!]
    \centering
    \includegraphics[width=\columnwidth]{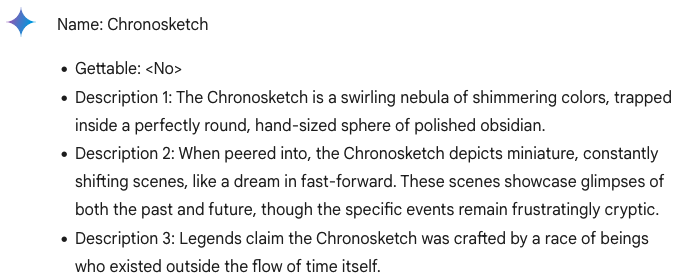}
    \caption{Gemini generates an item that fits the structured representation we propose.}
    \label{fig:example_generation}
\end{figure}

A possible example is shown in Fig.~\ref{fig:example_payador} and Fig.~\ref{fig:example_payador2}, where an LLM was used to generate the narration that a GM typically does to describe a scene~\cite{UniversalGameEngine}.
Another interesting gamemastering task is the creation of relevant items to be found by the players.
In Fig.~\ref{fig:example_generation} we show a successful preliminary experiment about that, prompting Gemini to generate an item that fits our representation (detailed in Table~\ref{tab:components}).

In other words, having a structured representation of the world connected to an LLM
may allow researchers to explore other possibilities.
In addition to those we just showed (narrate a scene and generate items), other examples may be
the application of many narrative generation approaches studied through the years~\cite{gervas2009computational,WANG2023126792}, or novel ideas born from research in content generation for games~\cite{sweetser2024large,gallotta2024large}.
We also think this can be beneficial for a clearer and unambiguous communication during the collaborative creative process between the automated GM and the players~\cite{benotti-blackburn-2021-grounding}, who might want to inspect and customize specific aspects of the fictional world or story.

\section{Conclusions}

In this paper we presented PAYADOR, an approach to the \textit{world-update} problem.
The main characteristic of our method is that we propose a change of focus: instead of modelling \textit{what} the player or items can do, model \textit{how} the fictional world can change.
To achieve this, we ground a Large Language Model to a minimal structured representation of the fictional world. 
The essence of this representation is to have specific attributes for only those details that are critical to keep the consistency of the fictional world; the rest of details are characterized using
\textit{independent descriptions}.
The code for the approach is available on GitHub.

We also showed and discussed some examples.
Having these LLMs grounded to a structured representation allows the system to run consistency checks, 
hence helping to prevent some of the unexpected behaviours 
reported by~\citeauthor{skill_check}~\shortcite{skill_check}.
Additionally,
while it is clear that its effectiveness is highly dependent on the LLM performance, other methods or models for common-sense reasoning may allow to improve it.
For instance, in this case we opted for a raw call to the Gemini API to keep the strategy simple, but additional machinery would make a remarkable improvement.

We think that one of the main strengths of PAYADOR is that it allows the user to understand what are the inconsistencies between the LLM and the actual world state.
Grounding neural models for dialogue is an increasing problem of interest~\cite{benotti-blackburn-2021-grounding} and we think it will remain being important in the future, even if stronger methods or models become available.
This kind of grounding may lead to improvements in Interactive Storytelling and digital RPG systems,
such as controlling the consistency of the fictional world, improving the communication between the user and the system during the co-creative process, or enhancing them with other related research in CC.

We hope this contribution helps to bridge the gap between classical and modern approaches in the NLP and CC fields, and to take another little step in this long path to model the rich mechanisms used by GMs to combine improvisation and planning to build their astonishingly beautiful fictional worlds.

\section{Acknowledgments}

This paper has been funded by ANII (Uruguayan Innovation and Research National Agency), Grant No. $POS\_NAC\_2022\_1\_173659$.

\section{Author Contributions}
SG is a MSc. student working on Interactive Storytelling. 
He carried out the preliminary experiments to study the feasability of the PAYADOR approach, and implemented the pipeline summarized in Fig.\ref{fig:architecture}.
PG (a collaborator), LC and GM (his advisors), contributed ideas, participated in the discussions, and revised the draft to make improvements for this version of the paper.

\bibliographystyle{iccc}
\bibliography{iccc}

\end{document}